\documentclass{ecai} 

\usepackage{latexsym}
\usepackage{amssymb}
\usepackage{amsmath}
\usepackage{amsthm}
\usepackage{booktabs}
\usepackage{enumitem}
\usepackage{graphicx}
\usepackage{color}
\usepackage{listings}
\usepackage{tikz}
\usetikzlibrary{positioning,shapes,arrows.meta}
\usepackage{tabularx}

\newcommand{\BibTeX}{B\kern-.05em{\sc i\kern-.025em b}\kern-.08em\TeX}

\lstnewenvironment{code}{}{}
\newcommand{\icode}[1]{\lstinline[basicstyle=\ttfamily]|#1|}

\begin{document}


\begin{frontmatter}


\paperid{1} 


\title{A User Study Evaluating Argumentative Explanations in Diagnostic Decision Support}


\author[A]{\fnms{Felix}~\snm{Liedeker}\orcid{0009-0006-2556-9430}\thanks{Corresponding Author. Email: fliedeker@techfak.uni-bielefeld.de.}}
\author[A]{\fnms{Olivia}~\snm{Sanchez-Graillet}\orcid{0000-0003-3483-265X}}
\author[C]{\fnms{Moana}~\snm{Seidler}}
\author[B]{\fnms{Christian}~\snm{Brandt}\orcid{0000-0001-8666-1640}}
\author[C]{\fnms{Jörg}~\snm{Wellmer}\orcid{0000-0003-2919-0496}}
\author[A]{\fnms{Philipp}~\snm{Cimiano}\orcid{0000-0002-4771-441X}}

\address[A]{Semantic Computing Group, CITEC, Bielefeld University, Bielefeld, Germany}
\address[B]{Bielefeld University, Medical School and University Medical Center OWL, Mara Hospital, Department of Epileptology, Bielefeld, Germany}
\address[C]{Ruhr-Epileptology, University Hospital Knappschaftskrankenhaus Bochum, Ruhr-University, Bochum, Germany}


\begin{abstract}
As the field of healthcare increasingly adopts artificial intelligence, it becomes important to understand which types of explanations increase transparency and empower users to develop confidence and trust in the predictions made by machine learning (ML) systems. 
In shared decision-making scenarios where doctors cooperate with ML systems to reach an appropriate decision, establishing mutual trust is crucial. In this paper, we explore different approaches to generating explanations in eXplainable AI (XAI) and make their underlying arguments explicit so that they can be evaluated by medical experts.
In particular, we present the findings of a user study conducted with physicians to investigate their perceptions of various types of AI-generated explanations in the context of diagnostic decision support. The study aims to identify the most effective and useful explanations that enhance the diagnostic process. 
In the study, medical doctors filled out a survey to assess different types of explanations. Further, an interview was carried out post-survey to gain qualitative insights on the requirements of explanations incorporated in diagnostic decision support. Overall, the insights gained from this study contribute to understanding the types of explanations that are most effective. 
\end{abstract}

\end{frontmatter}


\section{Introduction}

The use of ML models in medical decision support systems has been increasing recently~\cite{Khosravi2024_ArtificialIntelligenceDecisionMakingHealthcare}. Since false predictions of such systems can potentially have detrimental outcomes for humans, the need for explainability in ML-based decision support in medical systems has been widely recognised~\cite{Rudin2019_StopExplainingBlackBox,Antoniadi2021_CurrentChallengesFutureOpportunities}.
Explainability is a crucial element of trust~\cite{Horst2019OnTU}. Furthermore, explainability can play a key role in reducing biases and promoting a collaborative environment where AI and human decision-makers can work together seamlessly to enhance decision-making~\cite{Bertrand2022_HowCognitiveBiasesAffect}.

However, while research on eXplainable AI (XAI) has gained momentum, a precise definition of explainability and the requirements that make an explanation fit for purpose are lacking \cite{Speith2022_ReviewTaxonomiesExplainableArtificial, Arya2019_OneExplanationDoesNot}. This poses a significant challenge for developing AI systems because there is very limited guidance on how explanations should be designed to empower a human user to gain confidence in a prediction. 

Despite the acknowledgement of the diverse needs of stakeholders regarding XAI~\cite{Langer2021_WhatWeWantExplainable}, empirical evaluations of these approaches are relatively rare~\cite{Antoniadi2021_CurrentChallengesFutureOpportunities}. Addressing this lack of empirical evidence is essential though to validate and refine XAI techniques.

It is precisely this gap that we are addressing with our work by conducting a user study with medical experts to evaluate the assessment of different XAI approaches.
In particular, we rely on different methods to generate arguments that make the reasons explicit why a model made a certain prediction. For this, we define templates and instantiate these templates using standard XAI approaches, feature importance, and counterfactual explanations in particular. 

The research questions we would like to answer are:

\begin{itemize}
\item \textit{RQ1} How do users evaluate explanations generated by state-of-the-art XAI approaches?
\item \textit{RQ2} Are there differences in the way the different XAI approaches are evaluated?
\item \textit{RQ3} Would doctors rely on the XAI-generated explanations to explain their decision to a medical colleague?
\item \textit{RQ4} Which type of explanations would the experts use and how do they compare to XAI-generated explanations?
\end{itemize}

The above questions are designed to evaluate the level of confidence that medical users place into XAI-generated explanations and how they differ from the explanations they would give. 

We present the results of a user study involving eight medical experts in the field of neurology and epileptology from two hospitals. The user study was administered by way of a questionnaire relying on real patient cases. After receiving the questionnaire results, a debriefing was conducted to generate qualitative findings on how the explanations were perceived. 

The remainder of this paper is organized as follows: Section \ref{backgroud} provides some background on XAI methods. Section \ref{reletedWork} discusses related work on argumentation and XAI. Section \ref{methods} presents a real-world use case demonstrating the creation of argument-based explanations in natural language using the output of different XAI methods. This section also includes the evaluation of the generated explanations through a survey applied to expert users. In Section \ref{results}, we discuss the results of our methods and the user survey. Section \ref{conclusions} presents the conclusions and future perspectives of our work.

\section{Background} \label{backgroud}
The current state of XAI largely revolves around the explanation of black-box models by auxiliary methods - so-called post-hoc explanations~\cite{Arya2019_OneExplanationDoesNot}.
Beyond that, inherently interpretable white-box models that do not require auxiliary explanation methods are frequently used. Well-known examples of white-box models used (in the medical domain) are decision trees~\cite{Quinlan1986_InductionDecisionTrees} as well as Bayesian Networks (BN)~\cite{pearl2009causality}. 

In the area of post-hoc explainability, feature attribution methods and counterfactual explanations (CFs) are among the most frequently used methods. Feature attribution methods like SHAP (SHapley Additive exPlanations)~\cite{Lundberg2017_UnifiedApproachInterpretingModel} or LIME (Local Interpretable Model-agnostic Explanations)~\cite{ribeiro2016should} provide insights into the contributions of individual features to the predictions of ML models, enhancing interpretability by explaining model behaviour in a comprehensible manner. CFs provide indirect explanations through hypothetical, but similar counterexamples in the form of \textit{If X had been different, the prediction would have changed from P to Q}~\cite{jia2021enhancing}. CFs are especially popular because they resemble the way humans tend to explain decisions~\cite{Wachter2018_CounterfactualExplanationsOpeningBlack}.

So far, the XAI community has not devoted too much focus to the fact that explanations are supposed to increase confidence and empower humans. Studies evaluating the impact of explanations on users in terms of enhancing understanding, increasing confidence, or empowering users to make better decisions are rare, see e.g.~\cite{Meske2023_DesignPrinciplesUserInterfaces,Alufaisan_DoesExplainableArtificialIntelligence} for some exceptions. Meske and Bunde~\cite{Meske2023_DesignPrinciplesUserInterfaces} showed that XAI is able to increase perceived usefulness, ease of use, and intention to use of users. While results were not conclusive, the general potential to improve human accuracy in decision-making by leveraging XAI was shown by Alufaisan et al.~\cite{Alufaisan_DoesExplainableArtificialIntelligence}. A recent study by Hamm et al.~\cite{Hamm2023_ExplanationMattersExperimentalStudy} found that XAI, if resulting in a high degree of perceived explainability, can lead to positive effects on perceived usefulness and trust.  

Although XAI has become an important and active field of research, only limited attention has been paid to the question of what kind of explanations are actually most useful and under what circumstances. Approaches to overcome this shortcoming consist, for example, of investigating which general desiderata for explanations can be found~\cite{Miller2018_ExplanationArtificialIntelligenceInsights} or in examining various characteristics of explanations, such as the representation form of an explanation~\cite{Vilone2021_ClassificationExplainableArtificialIntelligence}. Furthermore, it has been shown that the generation process, i.e. the method used to generate an explanation, affects the user perception of such explanations~\cite{wang2021explanations}. This realm also includes investigations of explanation complexity, e.g. focusing on complexity imposed by constraints of different generation methods~\cite{Aechtner2022_ComparingUserPerceptionExplanations}, the presentation complexity of explanations~\cite{Huysmans2011_EmpiricalEvaluationComprehensibilityDecision}, or the overall explanation complexity due to explanation type and feature count within explanations~\cite{Liedeker2024_EmpiricalInvestigationUsersAssessment}. Results generally suggest the existence of a `sweet spot' between the complexity of an explanation and its value, quantified in terms of enhanced understandability or usefulness.

While a vast amount of different metrics have been proposed for automatically evaluating explanations, the gold standard for explanation evaluation is still the evaluation by humans~\cite{Doshi-Velez2017_RigorousScienceInterpretableMachine}, which underlines the importance of empirical user studies.

In the recent past, there has been a movement from the `classical', static, one-shot post-hoc explanations towards more human-centred and interactive XAI approaches. Rohlfing et al.~\cite{Rohlfing2021_ExplanationSocialPracticeConceptual} proposed a ``co-constructive'' approach to decision-making and explanation giving, aiming for a high level of interactivity and thus explainable decisions that are not dictated by an AI system but reached in close interaction between human and machine. Miller~\cite{Miller2023_ExplainableAIDeadLongb} argued for a paradigm shift towards ``evaluative AI'', proposing the superiority of machine-in-the-loop decision support which not only provides recommendations to a human user, but rather provides evidence supporting and challenging decisions, allowing the user to evaluate the evidence and taking a decision on this basis.

We believe that this offers a great opportunity, particularly in the area of decision support in medicine, to overcome users' reservations such as perceiving AI systems as a potential threat to their own occupation~\cite{Krittanawong2018_RiseArtificialIntelligenceUncertain} and raising awareness for the often underestimated risk of making errors in diagnostic decisions~\cite{Graber2022_Reaching95DecisionSupport} by not only keeping the ``doctor-in-the-loop''~\cite{Holzinger2019_CausabilityExplainabilityArtificialIntelligence}, but moving towards interactive, co-constructive decision support.

\section{Related Work} \label{reletedWork}

Explainability of AI systems aims to trace the steps that lead to decisions or conclusions. Explainability was relatively straightforward as early AI systems were logic-based, allowing transparency into how decisions were reached. However, as AI methods advanced, they became less transparent. XAI was then developed to provide transparency through explainability to these complex systems.
In tasks that require explanations, such as decision-making, justification, and dialogues across various fields of knowledge, argumentation can be very useful \cite{Vassiliades2021_ArgumentationExplainableArtificialIntelligence}.

Examples of argumentation frameworks that have been used to generate explanations include the Abstract Argumentation Framework (AAF) and the Assumption-Based Argumentation Framework (ABA) \cite{Fan2014, Fan_Toni_2015} which can be applied across different areas.

In recent years, XAI has integrated argumentation due to its strong capability to provide explanations. Argumentation can enable explainability by translating an AI system's decision into a step-by-step argumentation procedure that demonstrates the rationale behind the final decision \cite{Vassiliades2021_ArgumentationExplainableArtificialIntelligence}. For example, a given set of possible decisions can be mapped to a graphical representation with predefined attack/support elements that subsequently lead to the accepted decision, and show the steps followed.


In the field of medicine and healthcare, dealing with incomplete and contradictory information is necessary, and previous conclusions must be questioned when new evidence appears. Thus, argumentation is an important approach in developing applications that support decision-making by generating rationalized recommendations or explanations. There is a wide range of applications using argumentation in this field, including prognosis, diagnosis, planning and implementation of (pre)clinical trials, preparation of medical guidelines, and patient counselling. Argumentation-based techniques have been applied in the medical context since the mid-1980s, utilizing different representations and formalisms such as logic-based argumentation, abstract argumentation, argumentation schemes, and defeasible logic programming.

For example, Krause, Fox, and Judson \cite{krause1995} used the logic of argumentation (LA) \cite{krause1995LA}, which is a variant of first-order logic, in an application for qualitative reasoning in toxicology (StAR). The system provides arguments both for and against the compound being carcinogenic for each toxic alert. StAR showed that by using an argumentation approach, it was possible to offer valuable advice even when statistical information was not available.
Hunter and Williams \cite{hunter2012} established an argument-based framework based on Dung's model \cite{DUNG} for aggregating clinical trial evidence. This approach involves generating inductive arguments from evidence found in medical guidelines, and the superiority of interventions is determined based on specific preference criteria applied to these arguments. The method was assessed based on how well the results of the use cases aligned with those in published clinical guidelines.

Grando et al. \cite{grando2013} developed rules based on argument schemes and their critical questions to explain anomalous patient responses to treatments. They utilized the ASPIC platform \cite{Prakken2010}, an implementation of Dung’s model, to formalize and assess the acceptance status of the arguments. To evaluate their method, the authors compared the results provided by their method with a previous version that offered fewer explanations or justifications. Then, clinicians were asked to indicate their preferred presentation. Interestingly, in certain situations, clinicians preferred fewer explanations and textual over graphical ones.
More recently, defeasible reasoning \cite{Longo2013} and dialogue frameworks \cite{mann1988dialogue, Qurat-ul-ain2020} have been used in applications for the medical and healthcare field \cite{gomez2014, zielke2015}.

Nevertheless, there may be limitations when evaluating argumentation-based methods.
For instance, some methods are exploratory and lack comparison with similar approaches. Other methods focus on very specific applications, making it difficult to find comparable methods or results for comparison. Abstract representations are not always verbalized, which makes it difficult for humans to evaluate them. Most existing argumentation-based approaches do not offer a quantitative evaluation against a given gold standard of explanations due to the lack of gold standards for specific medical areas or goals. Finally, as Longo and Dondio \cite{Longo2014} pointed out, some approaches based on refutable reasoning and argumentation theory in the medical field rely on anecdotal evidence, which characterizes several use cases.

In spite of the fact that XAI and computational argumentation are conceptually close, it is only recently that work has emerged in a field denoted as \emph{argumentative XAI}, leveraging argumentative mechanisms to enhance explainability of ML systems (see \cite{Cyras2021_ArgumentativeXAISurvey,Vassiliades2021_ArgumentationExplainableArtificialIntelligence} for recent overviews). Following this new line of research, in this paper we combine computational argumentation and XAI, relying on XAI methods to instantiate Walton-like argumentative schemes \cite{walton2008}. We create such argumentative explanations by the verbalization in natural language of explanations generated by state-of-the-art XAI methods. As a first step, we evaluate the usefulness of such explanations in the context of a medical decision-making task.


\section{Methods}\label{methods}

To evaluate our AI-generated explanations in a real-world setting, we consider the use case of diagnosing transient loss of consciousness (TLOC)~\cite{Baumgartner2019_SynkopeEpileptischerOderPsychogener}. With TLOC, three main differential diagnoses need to be considered: i) epileptic seizures, ii) syncopes, and iii) psychogenic non-epileptic seizures (PNES)~\cite{Wardrope2018_DiagnosticCriteriaAidDifferential}. 

The differential diagnosis task is challenging since symptoms of TLOC mostly occurs spontaneously outside of diagnostic procedures. At the time of presentation to medical staff, patients might be asymptomatic again, making an accurate diagnosis based on the type of seizure challenging. 
In this situation, the key challenge for doctors is to inquire about the exact clinical course (the semiology) and framing conditions of the event including potentially facilitating or provoking factors. In particular, a differential diagnosis cannot be reliably determined solely based on the isolated presence or absence of a specific symptom~\cite{Oto2016_MisdiagnosisEpilepsyAppraisingRisks}. In case of a lack of distinct arguments for one particular type of TLOC (a gold standard), combinations of weaker arguments for one type, and exclusion criteria against other TLOC types, become the most important source of decision-making.

This state of affairs makes the diagnosis of epileptic seizures an ideal use case for AI-assisted decision support and therefore also for our user study. The three types of explanations utilised in our study are attribution-based explanations, counterfactual explanations (CFs), and explanations based on the exclusion principle, i.e. specifying why a certain diagnosis can be excluded from the set of possible hypotheses. Explanations and their generation process will be further detailed in the following.

\subsection{Data and Modelling}

The Ruhr-Epileptology at the University Hospital Knappschafts-krankenhaus Bochum, Germany, provided 32 anonymized and annotated transcripts of outpatient letters as the basis for our study. Annotations consist of the ground-truth diagnosis and the annotation of characteristics which, by presence or absence, support or oppose one of the three possible diagnoses (epileptic seizure, syncope, PNES). 
The dataset consists of ten patients with epileptic seizures, eleven patients with syncopes, and eleven patients with PNES.

\begin{figure}
\centering
\begin{tikzpicture}
\tikzset{every node/.style={draw,ellipse,inner sep=3pt,align=center,minimum width=60}}

\node (di) {\scriptsize Diagnosis};
\node[above=of di,xshift=-2.5cm,yshift=-0.6cm] (po) {\scriptsize Poor coordination};
\node[above=of di,xshift=2.5cm,yshift=-0.6cm] (fz) {\scriptsize Febrile seizures};
\node[below=of di,yshift=0cm] (oa) {\scriptsize Oral automatisms};
\node[below=of di,xshift=-3cm] (vs) {\scriptsize Violent shaking};
\node[below=of di,xshift=3cm] (mo) {\scriptsize ... 31 more ...};

\path[-{Latex[length=2mm,width=2mm]},shorten >=0.1cm] (po) edge (di);
\path[-{Latex[length=2mm,width=2mm]},shorten >=0.1cm] (fz) edge (di);

\path[-{Latex[length=2mm,width=2mm]},shorten >=0.1cm] (di) edge (vs);
\path[-{Latex[length=2mm,width=2mm]},shorten >=0.1cm] (di) edge (oa);
\path[-{Latex[length=2mm,width=2mm]},shorten >=0.1cm] (di) edge (mo);

\end{tikzpicture}
\caption{Three-layer Bayesian Network with the 36 most relevant variables determined by Wardrope et al.~\cite{Wardrope2020_MachineLearningDiagnosticDecision}.}
\label{fig:BN} 
\end{figure}
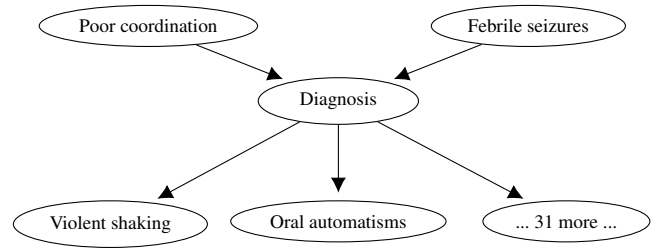

Since this small amount of data is insufficient to train a useful model, the data set was combined with data collected by Wardrope et al.~\cite{Wardrope2020_MachineLearningDiagnosticDecision}, who conducted a retrospective questionnaire study, collecting self- and witness reports from 300 patients (100 each with epilepsy, syncopes, and PNES). They initially collected 117 features and used a random forest approach to reduce the number of features to the 36 most important ones. With this data, we constructed a three-layer Bayesian Network (BN) disease model~\cite{Richens2020_ImprovingAccuracyMedicalDiagnosis}. We manually grouped variables into risk factors, diseases, and symptoms, encoded by binary nodes, indicating their presence or absence. The BN also allows modelling causal relationships between variables, i.e. risk factors causing diseases, which in turn are the cause of symptoms. The resulting structure of our BN is shown in Figure~\ref{fig:BN}. After defining the structure, parameters for the BN are learned from the aforementioned data. The BN serves as our ML model for predicting the diagnosis and determining the relevant feature used to construct the arguments for our study as it is explained in Section~\ref{sec:arg-creation}.

\subsection{XAI Methods}

The basis for generating our explanations are two XAI standard methods: LIME~\cite{ribeiro2016should} and CFs. LIME identifies the features that have the most significant impact on the model's output, thus highlighting the most important features for that particular prediction. The most influential features are determined by perturbing the input data, and observing the resulting changes in the prediction to highlight which features have the most influence on the prediction.

CFs offer insights by describing how altering specific input features could impact the model's prediction. These explanations provide similar but hypothetical counterexamples that help users understand how different variables can influence the outcome. We calculate CFs using the open-source Python library Alibi Explain~\cite{alibi_explain}.

However, the output of these methods is only feature-based. In the case of LIME, we get the most important features for a given prediction, i.e. diagnosis, and with the calculated CFs, we end up with features and their values that would change the prediction. These feature value pairs are used in conjunction with the annotations from the outpatient letters to construct argumentation-based explanations by instantiating different templates with these feature value pairs.  
In the following, we will describe this process and the underlying templates in more detail.

\subsection{Transforming the Output of XAI Methods into Arguments}\label{sec:arg-creation}

In this section, we describe the templates that are used to create argumentation-based explanations in natural language given the feature-based output of the used XAI methods.

\paragraph{Attribution-based explanation}

This type of explanation is based on the most important features that support the given diagnosis. The argument to support the prediction of \textit{diagnosis X} is created by instantiating the template:
\smallskip

\icode{Because feature1 is A and feature2 is B, the patient is most likely to have diagnosis X.}

We use this template and fill it with the two most important features identified by LIME. If these features are not among the relevant variables identified a priori for a certain diagnosis by our medical experts, we replace them with other less important features that fulfil this condition.

\paragraph{Counterfactual explanation}

While more complex than attribution-based arguments, CFs are still found to be easily understood by humans~\cite{Wachter2018_CounterfactualExplanationsOpeningBlack}. The argument to support the prediction of \textit{diagnosis X} is created by instantiating the template:
\smallskip



\icode{If feature1 was A and feature2 was B, the patient would most likely have diagnosis Y.}

\icode{If feature3 was C and feature4 was D, the patient would most likely have diagnosis Z.}
\smallskip

Where \textit{Y} and \textit{Z} are the two potential alternative diagnoses other than \textit{X}. We then generate a counterfactual for each of the possible alternative diagnostic outcomes. 

\paragraph{Exclusion principle explanation}

This type of explanation is based on the exclusion principle. The idea is to build an argument for why other possible diagnoses can be ruled out as possible hypotheses. The rationale behind this is to resemble the way doctors approach differential diagnosis tasks, i.e. by coming up with possible hypotheses based on the information available, refining those hypotheses, and ultimately eliminating hypotheses that are not sufficiently supported by the available 
evidence~\cite{Elstein2002_ClinicalProblemSolvingDiagnostic}. The argument to support the prediction of \textit{diagnosis X} is created by instantiating the template:
\smallskip

\icode{Because feature1 is A and feature2 is B, diagnosis Y can most likely be ruled out.}

\icode{Because feature3 is C and feature4 is D, diagnosis Z can most likely be ruled out.}
\smallskip

It is noteworthy that in the case of CFs and exclusion principle explanations, both explanations within the argument (targeting \textit{diagnosis Y} and \textit{Z}) could, in principle, share one common feature.

Individual explanations always contain two features, since this is a reasonable amount making explanations still easily comprehensible while not being overly simplistic. Different studies have indeed shown that having two features in an explanation is a reasonable choice~\cite{Keane2020_GoodCounterfactualsWhereFind,Liedeker2024_EmpiricalInvestigationUsersAssessment}.

Examples of template-generated explanations for a real case as shown in the user study are given in Table~\ref{tab:explanation-example} of the Appendix.

\subsection{Online User Study}

From the initial 32 outpatient letters, 10 cases (4 epileptic seizures, 3 syncopes, 3 PNES) were randomly selected for inclusion in our user study. Based on these letters, slightly shortened and anonymized case reports have been compiled. The user study was structured as follows:

\paragraph{Participant's Background}

Using a first set of four questions, we captured information about the participants, including their age, sex and experience in diagnosing epileptic seizures as well as experience of dealing with AI systems. Questions about users' experience were answered on a 5-point Likert scale from 1 (no experience) to 5 (expert level).

\paragraph{Case Introduction}

In the user study, ten cases of real patients were described using a short case report\footnote{An example case report is provided in the Appendix}. These cases were constructed based on the anonymized transcripts of outpatient letters provided by our partners.\\

After participants were presented with a case report, they were asked about the following specific aspects: 

\paragraph{Own diagnosis}

Participants were asked to provide their own estimate of the diagnosis of the patient given the textual description. They had the possibility to select one option from the three possible diagnoses: epileptic seizure, syncope or PNES. Further, they were asked to rate the statement \textit{`I am sure about the diagnosis.'} on a 5-point Likert scale from 1 (does not apply at all) to 5 (fully applies) to assess the user's confidence in the diagnosis. In addition, participants were prompted to select the most likely sub-classification of the diagnosis (if applicable). All possible sub-classifications are summarised in Table~\ref{tab:sub-classification}.

\begin{table}[h]
\caption{Possible sub-classifications of diagnoses. Note: The diagnosis of PNES has no sub-classifications.}
\centering
\begin{tabular}{l@{\hspace{8mm}}} 
\toprule
Epileptic: focal\\
Epileptic: multifocal\\
Epileptic: generalized\\
Epileptic: focal to bilateral tonic-clonic\\
Epileptic: other\\
Syncope: cardiogenic\\
Syncope: vaso-vagal\\
Unclear\\
\bottomrule
\end{tabular}
\label{tab:sub-classification}
\end{table}

Before assessing AI-generated explanations, users were asked to provide their own explanations about their proposed diagnosis. This was done to compare their explanations with the ones provided by the XAI system.
The goal was to determine if there was sufficient objectivity in the task and cross-expert agreement. 

\paragraph{Assessment of explanations}

For each case, our three different argumentative explanations are shown one after one to the user. Experts were asked to rate the following statements on a Likert scale from 1 (does not apply at all) to 5 (fully applies), in order to assess the perceived value of each explanation.

\begin{itemize}
\item \textit{The explanation is comprehensible}.
\item \textit{The explanation is plausible}.
\item \textit{The explanation is complete and contains the most important features to justify the diagnosis}.
\item \textit{I would use a similar explanation to explain the facts to a colleague}.
\end{itemize}

The study was conducted online in an anonymous fashion via a link that was sent to all participants. Participation in the study took around one and a half hours.

Eight physicians were recruited from the Department of Epileptology at Mara Hospital, Bielefeld University and the Ruhr-Epileptology, University Hospital Knappschaftskrankenhaus Bochum, as participants for our user study. Four participants (50\%) were male and four participants (50\%) were female. The age of the participants ranged from 27 to 54 years (mean = 42.29, standard deviation (SD) = 9.53). Participants reported a mean experience with the differential diagnosis of epileptic seizures of 4 (SD = 1) on a scale from 1 (No experience) to 5 (Expert level), whereas the reported experience in dealing with artificial intelligence was 1.71 (SD = 0.48) on the same scale.

\subsection{Follow-up Interviews}

After the online study was completed and we have analysed the results, we conducted semi-structured interviews with our participants to gather qualitative data on explanation preferences or requirements in order to resolve questions arising from the online study. Interviews were done online via Zoom, taking around 45 minutes per participant. In particular, we went through the following topics with the participants:

\paragraph{Discussion of results from the user study}

At the beginning of each interview, the results of the study, cf. Section \ref{sec:results}, were presented and discussed.

\paragraph{AI-generated explanations vs. user-provided explanations} To further elaborate on \textit{RQ3} and \textit{RQ4}, i.e. the question of the reliability of AI-generated explanations to explain a case to a colleague and to gain quantitative data on what, according to medical experts, constitutes a good explanation, the experts were specifically asked why they think the discrepancy between the reliability rating and user-provided explanations with similar patterns exists, as described in Section~\ref{sec:results}. Also, the question of how the existing, AI-generated explanations should be improved to meet the users' expectations was discussed in the follow-up interviews. 

\paragraph{Related topics}In addition to the previously described topics directly aimed at answering our research questions, the following topics have been addressed in the follow-up interviews: 

\begin{itemize}
    \item Importance of interactive or co-constructive explanations.
    \item Other explanation formats beyond textual representations, e.g. diagrams.
    \item Major challenges arising from implementing AI decision support in clinical practice.
    \item General remarks or questions from the users.
\end{itemize}


\section{Results}\label{results}
\label{sec:results}
In the following, we present the results of our study and the results obtained from the conducted follow-up interviews.

In order to check the inter-expert agreement on the diagnoses as well as the self-assessment of the experts, we first looked at the results of the user-provided diagnoses and the reported confidence to compare them among each other and with the ground truth diagnosis provided by our cooperation partners.

The overall certainty of diagnoses across all experts and cases is very high with a mean score of 4.13 (SD = 1.04). This is also reflected in the fact that, in 8 out of 10 cases, all participants agreed on the correct diagnosis. Only the sub-classification of diagnoses revealed some deviations. For example, in one case participants selected `focal', `multi-focal', and `focal to bilateral tonic-clonic' as their sub-classifications for an epileptic seizure. Note that this could also be due to the fact that case reports were a bit shortened and the physicians may not have been able to ask questions about the patients, as it would typically occur in a real diagnostic scenario.

In terms of diagnostic certainty, only two cases showed substantially worse results than the other cases: Case 6 (certainty 2.5 (SD = 0.76) and Case 9 with a certainty of 2.8 (SD = 1.07). However, in both cases, experts agreed on the diagnosis (`Epileptic: focal') except for two participants who gave no answer on the correct diagnosis.

\begin{figure}[h]
\centering
\includegraphics[width=7cm]{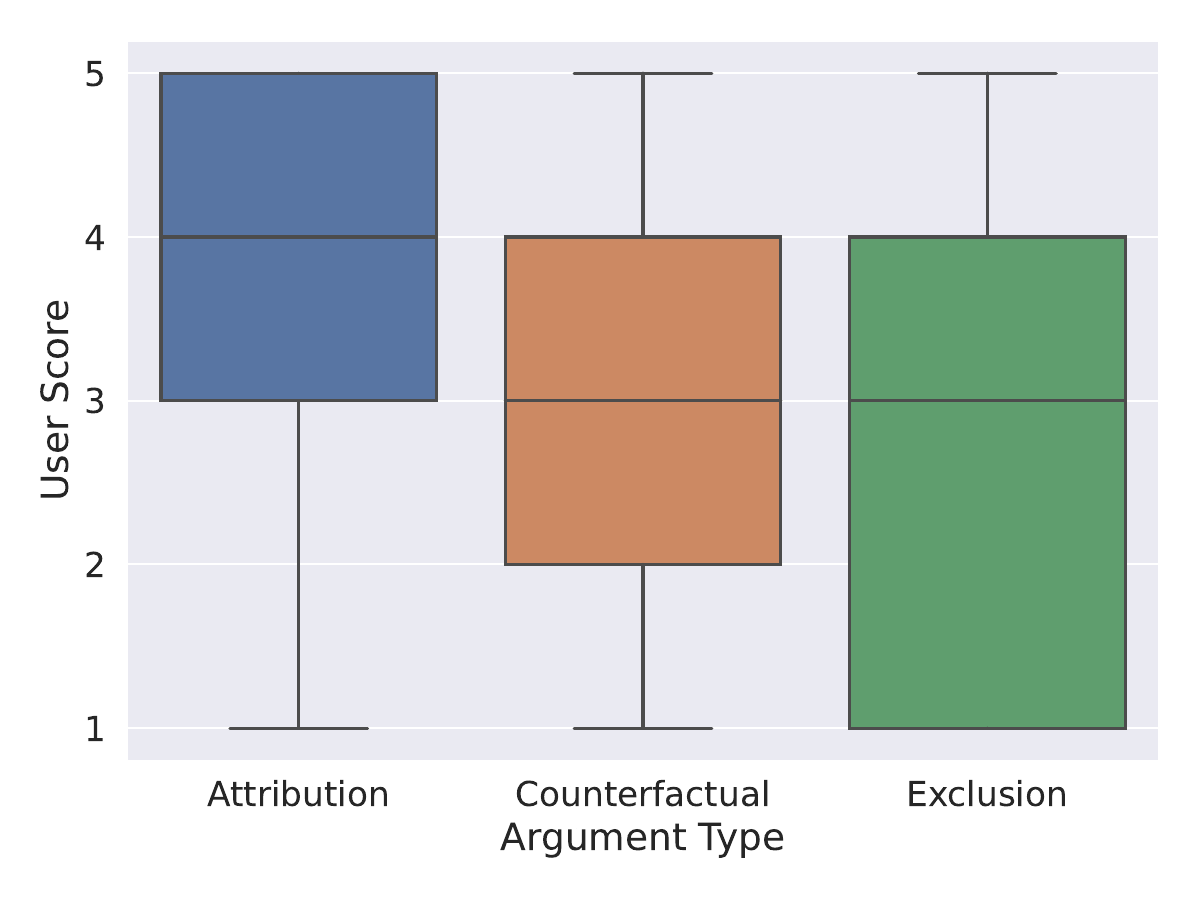}
\caption{User ratings on the \textbf{comprehensibility} of the three generated argument types.}
\label{fig:comprehensible}
\end{figure}

\begin{figure}[h]
\centering
\includegraphics[width=7cm]{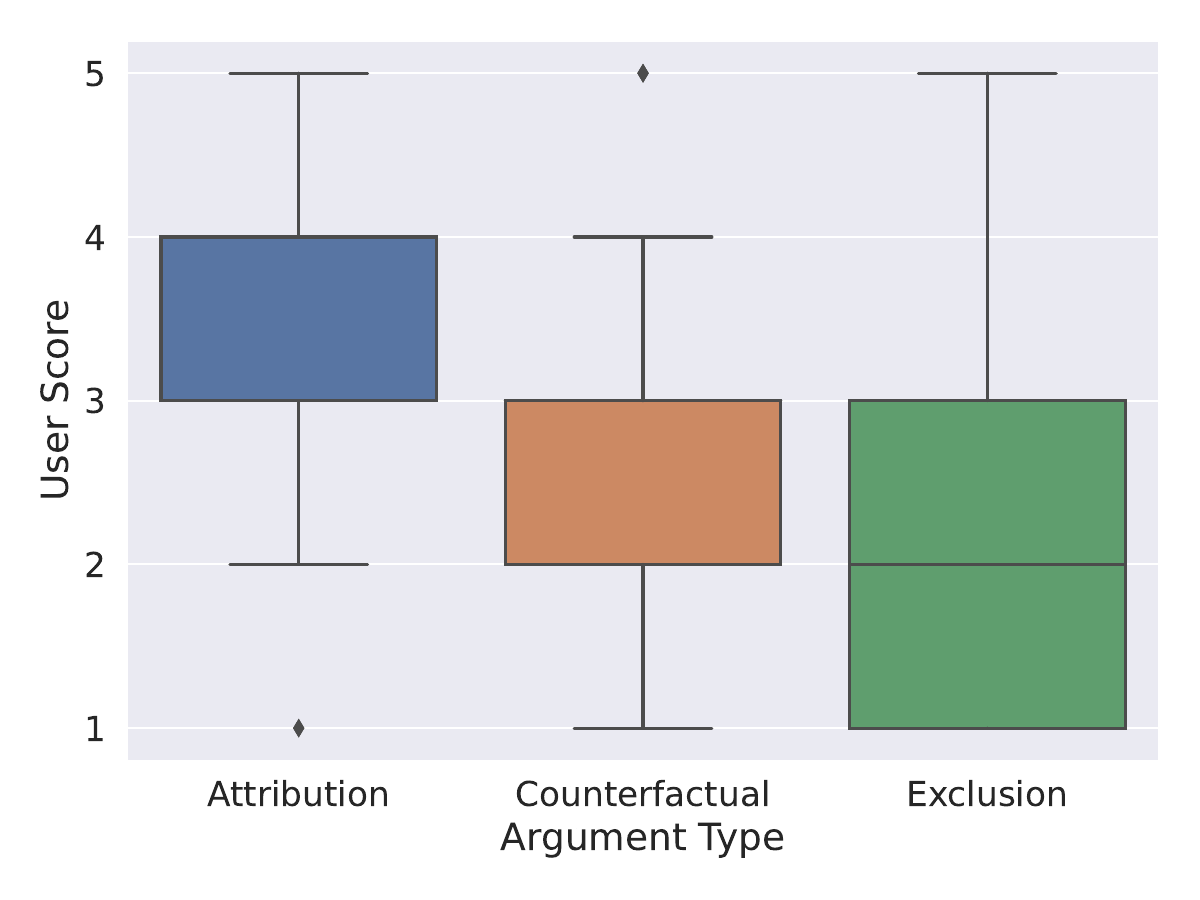}
\caption{User ratings on the \textbf{plausibility} of the three generated argument types.}
\label{fig:plausible}
\end{figure}

In the following, results are visualized with boxplots and it must be noted, that in some cases (e.g. attribution and counterfactual arguments in Figure~\ref{fig:plausible}), the median overlaps with one of the quartiles and is therefore not recognizable in the plot. 

Figure~\ref{fig:comprehensible} shows the results on user comprehensibility of the different argument types. It can be seen that the comprehensibility is always at least moderate to good (median > 3). Users perceived the attribution-based arguments by far as the most comprehensible argument - this applies to both the median value and the distribution of responses. For both CFs and attribution-based explanations, the median value is significantly lower and answers are more distributed on the rating scale.

Concerning the plausibility of explanations, which is depicted in Figure~\ref{fig:plausible}, a similar trend can be observed: Attribution-based arguments again achieved the best rankings on plausibility. Compared to the comprehensibility of the arguments, CFs and exclusion-based arguments fall further behind the attribution-based arguments and reported user scores are more concentrated (on the lower spectrum of the scale).

\begin{figure}[h]
\centering
\includegraphics[width=7cm]{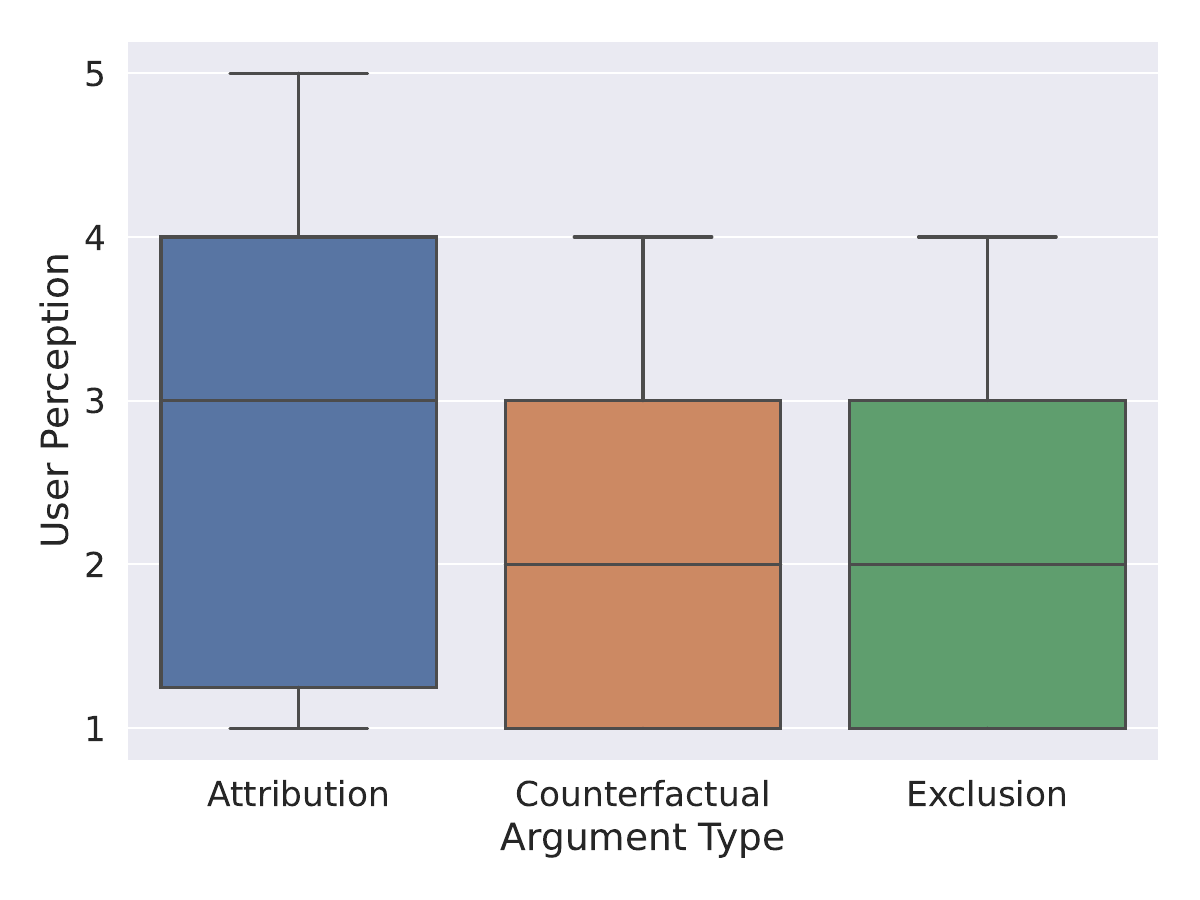}
\caption{User ratings on \textbf{completeness} of the three generated argument types.}
\label{fig:complete}
\end{figure}

The completeness of all argument types is perceived as rather low, or moderate in the case of attribution-based arguments as shown in Figure~\ref{fig:complete}. 
The differences observed in this property are quite surprising, considering that all explanations contain the same number of features. Additionally, both the counterfactual and exclusion-based arguments include alternate diagnoses. Therefore, one would expect these to be perceived as more complete arguments. However, as shown in Figure~\ref{fig:complete}, this is not the case.

To summarise the results of how users evaluated explanations generated by state-of-the-art XAI approaches (\textit{RQ1}), it can be said that arguments generated by state-of-the-art methods are perceived moderately good or good, depending on the evaluated dimension under consideration.

\begin{figure}[h]
\centering
\includegraphics[width=7cm]{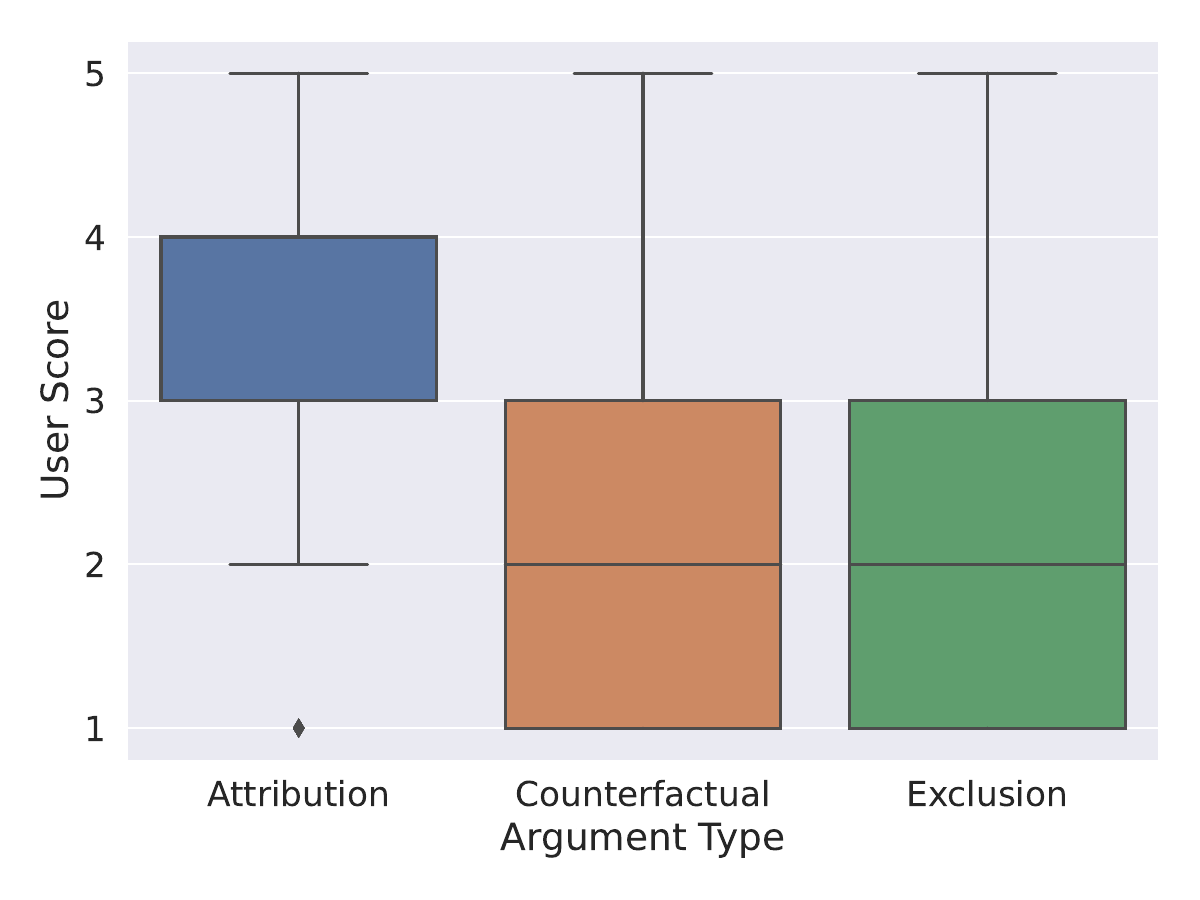}
\caption{User ratings along \textbf{all dimensions combined} for each argument type.}
\label{fig:combined}
\end{figure}

The differences between the individual argument types can be very marked sometimes, e.g. when comparing the estimated plausibility of an explanation to the question of whether experts would use similar explanations in a discussion with a colleague. On the other hand, differences in comprehensibility and completeness are less significant, as all argument types receive similar evaluations in these aspects. These insights also answer \textit{RQ2} and the question about differences in the evaluation of the different types of arguments. Figure \ref{fig:combined} shows the combined results for all evaluated dimensions per argument type. It can be seen that counterfactual-based arguments and arguments based on the exclusion principle are very similar with respect to the median score in the combined analysis, while the simpler arguments based on feature attribution achieved better results.

\begin{figure}[h]
\centering
\includegraphics[width=7cm]{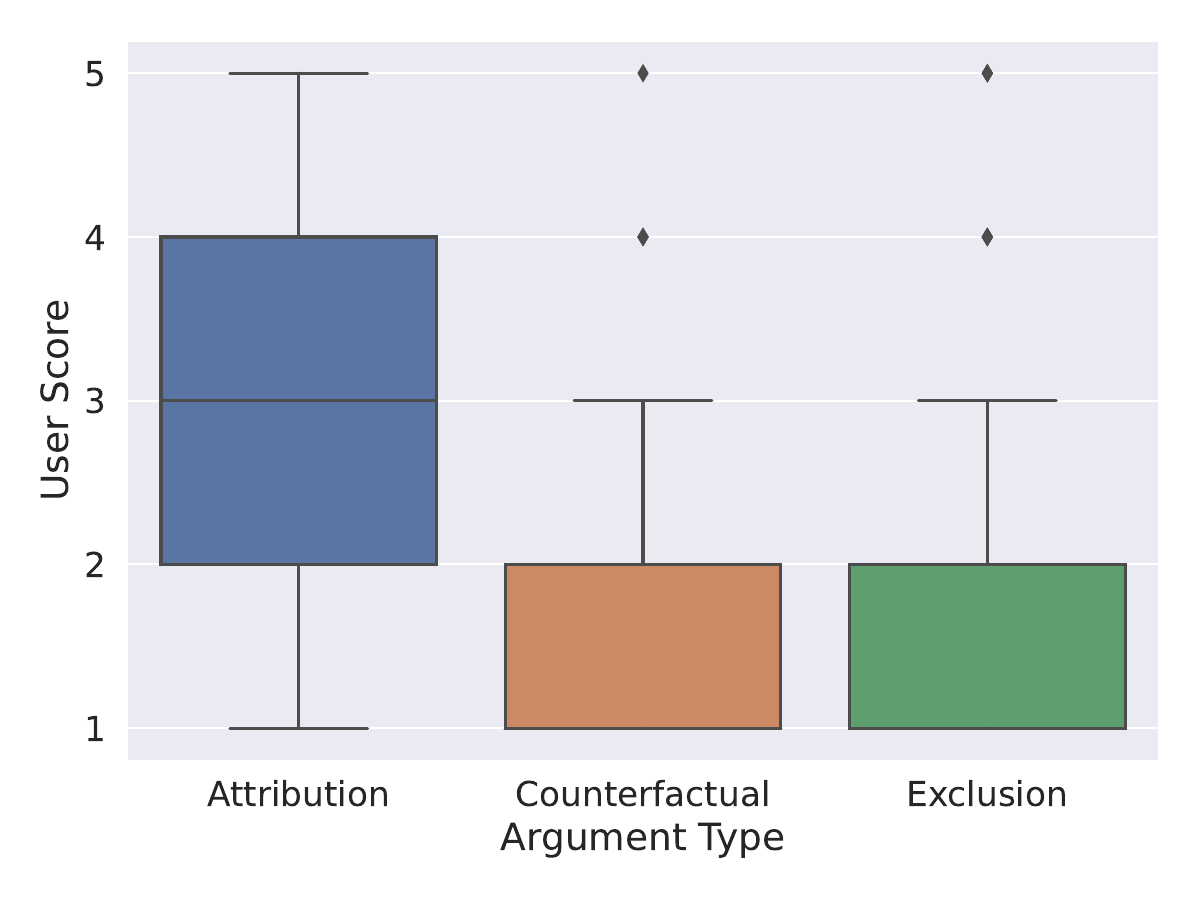}
\caption{Ratings for the statement \textit{\textbf{``I would use a similar explanation to explain the facts to a colleague''}} for our three argument types.}
\label{fig:colleague}
\end{figure}

When users were asked whether they would use a similar explanation to the AI-generated one to explain a case to a colleague, most experts assigned rather low scores for all argument types. Although some experts (see outliers in Figure~\ref{fig:colleague}) would use all three types, including CFs and exclusion-based explanations to explain a case to a colleague, in most cases they would not use XAI-generated explanations to explain their decision to a colleague.
This finding seems to stand in opposition to the assessment of explanations along the dimensions of plausibility, comprehensibility, and completeness. 

We thus decided to investigate the reasons for this discrepancy in qualitative follow-up interviews (see below). 

To address \textit{RQ4}, which is the question of what types of explanations experts would use and how they compare to XAI-generated explanations, we analysed the responses provided by participants when they were asked to give a concise explanation of the diagnosis. We identified distinct types or patterns in the provided explanations and summarised them in Table~\ref{tab:user-explanations}, ordering them according to the number of occurrences.

Interestingly, the exclusion of other diagnoses is a pattern that was frequently encountered in the user-provided explanations. However,  experts gave low scores to this kind of explanation. This also applies to the listing of symptoms which was the third most frequent pattern found in the users' explanations. This pattern is similar to our attribution-based arguments. Although this type of explanation was evaluated better, based on the analysis of user explanations, one might have expected a higher ranking.

\begin{table}[h]
\caption{Explanation patterns found in user-provided explanations.}
\centering
\begin{tabular}{l@{\hspace{8mm}}} 
\toprule
Prototype  \\
Exclusion of other diagnoses \\
List of symptoms \\
Prototype \& Exclusion of other diagnoses \\
Prototype \& List of symptoms \\
List of symptoms + Exclusion of other diagnoses\\
\bottomrule
\end{tabular}
\label{tab:user-explanations}
\end{table}

Another interesting finding is that experts often construct an explanation by combining different patterns. For example, they may combine a list of symptoms that support a diagnosis with the mention that this case is a good prototype of the diagnosis:

\icode{Typical semiology for temporal lobe seizures of the linguistic dominant hemisphere and corresponding epilepsy-typical potentials on the left temporal side.}\footnote{Translated from German.}

\smallskip

Another combination encountered multiple times is the exclusion of other diagnoses with mentioning the prototype character of the considered case:

\icode{Typical anamnesis and description of the seizure, no indications for differential diagnoses.}\footnotemark[\value{footnote}]

\paragraph{What is missing in XAI-generated explanations?}

To address the discrepancy between expert's rating along dimensions of plausibility, comprehensibility, and completeness, and the fact that our users would rather not use an AI-generated explanation to explain a case to a colleague; in addition to further answering \textit{RQ3}, this topic was specifically addressed in the follow-up interviews. 

When asked about this matter, participants stated that they had not received any information about the quality and reliability of the generated explanations and the uncertainty of the prediction. This made them believe that they might not be able to trust the provided argumentative explanations and, as a result, they would not use them in practice. Participants indicated that information on both the quality of the explanation and the prediction of the AI system would be important. 

Besides this regard, the role of personal preferences and the fact that it is difficult to develop a universal solution is also evident when it comes to additional information that could be included in the system, such as graphics, statistics on the patient population, etc.. While some participants would like to receive more information, others expressed concern that too much information could lead to cognitive overload or less intent to use a support tool.

Furthermore, participants found it problematic that the same types of argument were generated and displayed for all cases, which contradicted their intuition that explanations should vary depending on the complexity or ambiguity of a case. In addition to the suggestion to adapt explanations to the complexity of the case, the suggestion was also made to take into account the experience of the doctor and to provide less experienced doctors with more detailed explanations.

\paragraph{Interesting findings from the follow-up interviews not directly related to our research questions}

The following are some insights we gained from the interviews that are not directly related to our research questions, but are nonetheless worth mentioning:

\begin{itemize}
    \item General criticism of the test setup. The constrained design of the study does not correspond to a realistic clinical setting in which the expert would rely on the help of an AI system.
    \item Participants have expressed that it might be difficult to convince doctors to use a support tool in general.
    \item The presented study design might also be useful in teaching scenarios or for conducting quality assurance measures within the hospital setting.
\end{itemize}


\section{Conclusion and Outlook}\label{conclusions}

In this paper, we have presented an empirical study to evaluate the usefulness of argumentative explanations generated by state-of-the-art XAI methods with medical users. The application context was in the field of diagnostic support and we considered explaining diagnostic decisions for transient loss of consciousness. Towards this goal, we have developed a simple template-based approach to transform the output of feature attribution methods such as LIME as well as counterfactual approaches into natural language arguments. 

In a user study, the generated arguments were evaluated along four dimensions: plausibility, comprehensibility, completeness and applicability to expert explanations. 
While we found that all explanations score reasonably well along the first three dimensions, the attribution-based methods received the best overall or combined user evaluations. Depending on the considered dimensions, differences between explanation methods became very distinct: In terms of plausibility and comprehensibility, attribution-based arguments were evaluated significantly better than counterfactual or exclusion-based arguments.
A surprising result was that, even though the overall assessment of the generated explanations was moderately good, the experts were reluctant to use them to explain their decisions to their colleagues. In a follow-up qualitative study, we could identify two main reasons for this discrepancy: 

The lack of information regarding the quality and reliability of AI-generated explanations, as well as the uncertainty of the predictions, which led to a general lack of trust in the AI system.
Furthermore, the lack of specificity in the explanations generated for a particular case was identified as a reason for experts' reluctance to use them to explain a case to a colleague. Since we were able to determine patterns in the user-supplied explanations that are very similar to the patterns we used to generate arguments, we believe that improving the specificity of explanations could lead to better AI-generated arguments.

In terms of limitations, it is important to mention that our population was very small. Yet, they are all experienced physicians who have scarce availability. Given their obvious domain expertise, we regard their answers as representative despite the smaller population.

As a next step, we intend to conduct a further study in which we leverage the insights from the current study and construct arguments that have a higher specificity for the given case, combining different XAI methods to generate more comprehensive argumentative explanations.



\begin{ack}
This research was funded by the \textit{Deutsche Forschungsgemeinschaft} (DFG, German Research Foundation): TRR 318/1 2021 – 438445824.
The research of Olivia Sanchez-Graillet was funded by the project RecomRatio (No. 376059226) as part of the DFG Priority Program “Robust Argumentation Machines (RATIO)” (SPP-1999).\\

We gratefully acknowledge and thank the physicians who participated in our user study.
\end{ack}



\bibliography{mybibfile}


\section*{Appendix}
\label{sec:appendix}

In the following, we show the translation into English of a German case report as it was part of the study. Case reports consist of three sections: Anamnesis, semiology, and examination results:\\

\begin{itshape}
Medical history: The patient had an event of the semiology specified below. A similar event had occurred approximately one year earlier. The patient's medical history is unremarkable for typical epilepsy-predisposing factors, the family history is also negative. Without previous internal diseases. In a previously performed EEG, potential sequences were classified as suspicious for sharp-slow-wave and thus as possibly typical for epilepsy.

Semiology: While sitting playing Monopoly, dizziness, then stood up, felt weak and dizzy on the way to the toilet, knees weak, then blacked out, loss of consciousness, and fall. According to anamnestic information from others, the duration of unconsciousness is significantly less than 20 seconds, with no motor manifestations. Quickly re-orientated.
Earlier event: comparable onset, but in the further course the patient also felt a cramping of the fingers and hyperventilated.
Beyond this, there is no indication of the presence of simple-partial, complex-partial, or secondary-generalised seizures or of myoclonus or absences.

EEG: Well-defined, very regular alpha EEG around 9/sec, no general changes. No foci, no epilepsy-typical potentials. Under repeated vigilance fluctuation, however, frequent groups of generalised, frontally accentuated high-tension alpha and theta waves, but in each case without evidence of clear epilepsy-typical potentials (spike or spike wave). Taking into account the temporal association with phases of vigilance reduction, these are hypnagogic theta groups. The EEG found again in retrospect can also be interpreted as an EEG with hypnagogic theta groups. The single detection of an isolated right-temporal delta wave is not sufficient for the diagnosis of a focal finding.
\end{itshape}

\begin{table}[h]
\caption{Examples of AI-generated and two user-provided explanations for the same case as it was present in the user study. AB: Attribution-based explanation, CF: Counterfactual explanation, EB: Exclusion principle explanation. All text was translated from German.}
    \begin{tabularx}{\linewidth}{lX}
    \toprule
        AB & Because the episodes occur exclusively while standing or walking, and lying down can alleviate the symptoms, the patient very likely has syncope.\\
        CF & If the episodes had clear provoking factors and the course of the episodes fluctuated significantly, then the most likely diagnosis would be a psychogenic seizure.\newline If the EEG were to show epilepsy-specific potentials and the episodes did not exclusively start from standing or walking, then the most likely diagnosis would be an epileptic seizure.\\
        EB & Because the episodes do not have clear provoking factors and the EEG does not show epilepsy-specific potentials, an epileptic seizure can be ruled out with high probability.\newline Because no psychiatric preconditions are known and no uncoordinated movement patterns are present, a psychogenic seizure can be ruled out with high probability.\\
        \midrule
        P2 & Provoking factors, the possibility of interruption by lying down, only while standing and walking.\\
        P3 & Exclusively while standing and walking, can be mitigated by sitting down.\\       
        \bottomrule
    \end{tabularx}    
    \label{tab:explanation-example}
\end{table}

\end{document}